\documentclass{article}
\ifdefined\pdfinfoomitdate
\fi
\ifdefined\pdftrailerid
  \pdftrailerid{}
\fi
\PassOptionsToPackage{numbers,sort&compress}{natbib}
\usepackage[eandd, preprint]{neurips_2026}
\usepackage[utf8]{inputenc}
\usepackage[T1]{fontenc}
\usepackage{amsmath,amsfonts,amssymb}
\usepackage{booktabs}
\usepackage{graphicx}
\usepackage{float}
\usepackage{multirow}
\usepackage{array}
\usepackage{enumitem}
\usepackage{microtype}
\usepackage{url}
\usepackage{xspace}
\usepackage{xcolor}
\usepackage{hyperref}
\hypersetup{
  hidelinks,
  pdfauthor={Wenyu Zhang, Guoliang You, Tianlun, Haotian Zhao, Tianshu Zhu, Haoran Wang, Xiaoxuan Tang, Mingyang Dai, Jingnan Gu, Daxiang Dong, Jianmin Wu},
  pdftitle={},
  pdfsubject={},
  pdfkeywords={},
  pdfcreator={},
  pdfproducer={}
}
\newcommand{\benchmark}{WebGameBench\xspace}
\newcommand{\specname}{Structured WebGame Specification\xspace}
\newcommand{\excellent}{\textsc{Excellent}}
\newcommand{\usablelabel}{\textsc{Usable}}
\newcommand{\unusable}{\textsc{Unusable}}

\title{WebGameBench: Requirement-to-Application Evaluation for Coding Agents via Browser-Native Games}
\author{
Wenyu Zhang\textsuperscript{1,*} \quad
Guoliang You\textsuperscript{2,*} \quad
Tianlun\textsuperscript{1} \quad
Haotian Zhao\textsuperscript{1} \\
Tianshu Zhu\textsuperscript{1} \quad
Haoran Wang\textsuperscript{1} \quad
Xiaoxuan Tang\textsuperscript{1} \quad
Mingyang Dai\textsuperscript{1} \\
Jingnan Gu\textsuperscript{1,\textdagger} \quad
Daxiang Dong\textsuperscript{1,\textdagger} \quad
Jianmin Wu\textsuperscript{1,\textdagger} \\
\normalfont\textsuperscript{1}Baidu \quad
\textsuperscript{2}University of Science and Technology of China \\
\normalfont\scriptsize\texttt{\{zhangwenyu08,tianlun,zhaohaotian02,zhutianshu,wanghaoran11\}@baidu.com} \\
\normalfont\scriptsize\texttt{\{tangxiaoxuan02,daimingyang,gujingnan,dongdaxiang,wujianmin\}@baidu.com} \\
\normalfont\scriptsize\texttt{glyou@mail.ustc.edu.cn} \\
\normalfont\scriptsize * Equal contribution. \quad
\textdagger{} Corresponding authors.
}

\begin{document}
\maketitle

\begin{abstract}
Coding agents are increasingly used as application builders, yet many evaluations still focus on source code, repository-level tests, or intermediate traces rather than the delivered application. We introduce \benchmark{}, a requirement-to-application benchmark that evaluates whether coding agents can turn a frozen \specname{} into a browser-accessible game. Browser-native games provide a compact but behavior-dense testbed: even simple games require coordinated input handling, spatial mapping, rule execution, state transitions, terminal conditions, restart behavior, and visible feedback. In \benchmark{}, each generated artifact is built, served, and exposed as a browser-accessible application under a unified deployment protocol. A runtime evaluator then interacts with the delivered game in a real browser and assigns a three-way label: \excellent{}, \usablelabel{}, or \unusable{}. On a human-reviewed subset, the runtime label is broadly aligned with human gameplay review under the Usable-rate criterion. Across 111 tasks, 12 coding agents, and 14 evaluation configurations, \benchmark{} separates current systems: the best configuration reaches a 76.9\% usable rate but only a 20.2\% excellent rate. This gap shows that crossing the minimum playable-delivery threshold is still far from complete requirement satisfaction. \textbf{To our knowledge, \benchmark{} is the first requirement-to-application benchmark for browser-native game delivery that validates delivered-application runtime labels against independent human gameplay review under the Usable-rate criterion.}
\end{abstract}

\section{Introduction}

As coding agents move from local code generation toward building applications from requirements, evaluation should move beyond code correctness and ask whether a model can understand a requirement, implement a system, deploy it locally, and leave users with an operable application. Existing benchmarks cover several important slices of this capability chain: code benchmarks such as HumanEval \citep{HumanEval-arXiv21} and LiveCodeBench \citep{LiveCodeBench-ICLR25} focus on function- or program-level correctness; software-engineering benchmarks such as SWE-bench \citep{SWE-bench-ICLR24} and Terminal-Bench \citep{Terminal-Bench-arXiv26} focus on long-horizon work in existing repositories or terminal environments; and web and computer-use benchmarks such as WebArena \citep{WebArena-ICLR24} and OSWorld \citep{OSWorld-NeurIPS24DB} evaluate task completion in existing interactive environments.

However, these directions do not directly evaluate the full path from a requirement to a delivered interactive application. They typically score code, patches, terminal outcomes, or actions in existing environments, whereas a delivered application may install, build, and load while still failing to satisfy the required runtime behavior. Evaluation for this setting should therefore examine whether the application itself behaves according to the specification during real interaction.

We call this setting \emph{requirement-to-application evaluation}. Each task provides a frozen structured requirement; the coding agent must synthesize a source artifact and deliver an application that users can operate through its interface. The evaluated unit is not an intermediate reasoning trace, isolated code fragment, repository diff, or action sequence in an existing environment, but whether the delivered application satisfies the requirement at runtime. This boundary matters because buildability and loadability are not application usability: a project may install, build, and render a page while still violating the specification in input control, spatial mapping, rule execution, state transitions, visible feedback, terminal conditions, or restart flow.

To instantiate this setting reproducibly, a benchmark needs an application substrate that reduces dependence on external services, private assets, and complex system configuration, while still supporting lightweight deployment, a unified access point, and user-level automated interaction. Browser-native web applications satisfy these requirements. In our instantiation, the generated application is deployed locally and exposed through a browser-accessible URL, which cleanly separates application synthesis from interactive runtime evaluation. This shifts assessment from source-level plausibility to the behavior of the delivered application in the browser.

\begin{figure}[t]
  \centering
  \includegraphics[width=0.78\linewidth]{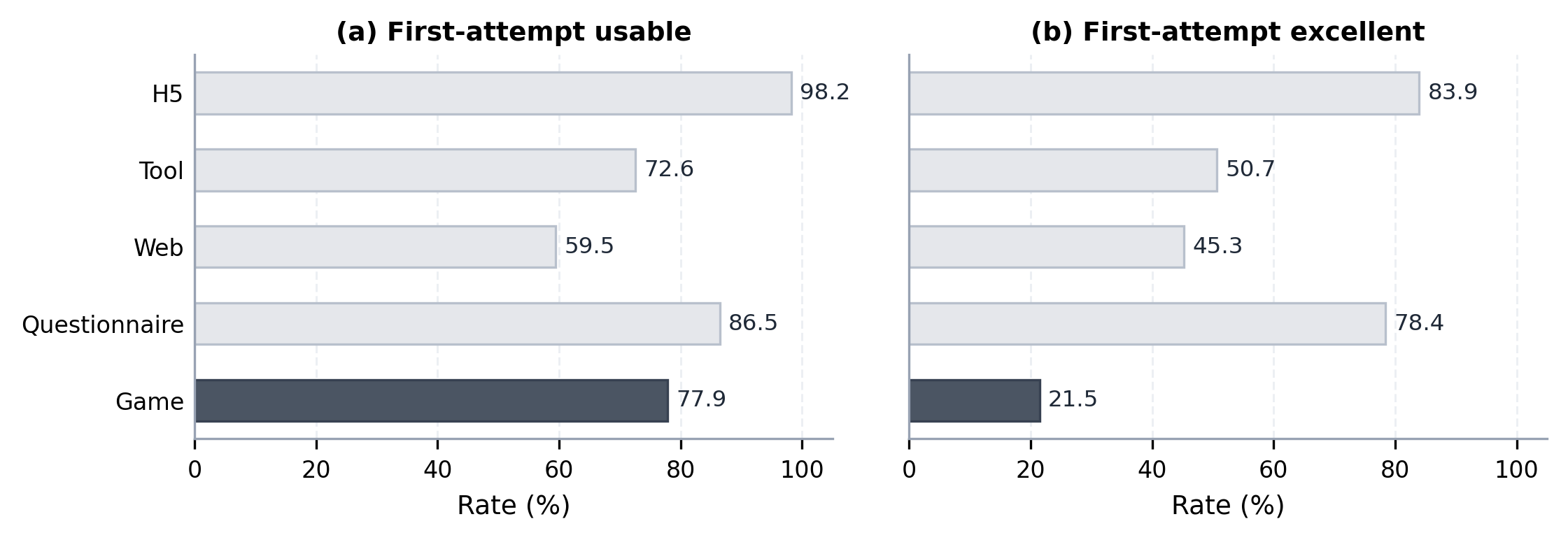}
  \caption{Pilot study on browser-native artifacts. We use `Opus 4.6' to compare H5, Tool, Web, Questionnaire, and Game tasks under the same generation and evaluation process. Usable rate counts artifacts labeled \excellent{} or \usablelabel{}; excellent rate counts artifacts labeled \excellent{}. The result is used only to determine the benchmark scope, not for main model comparison.}
  \label{fig:toy}
\end{figure}

Among browser-native artifacts, we use games as the primary testbed. This choice is not because game generation is the endpoint of the research problem, but because games compress dense dynamic behavior into small specifications: input response, spatial relationships, collision or hit detection, score and resource updates, state machines, win/loss conditions, restart logic, and visual feedback. These behaviors require the model not only to generate a loadable interface, but also to preserve the runtime dynamics specified by the requirement. To check that games are a practical substrate rather than only an intuitively appealing one, we conduct a pilot study. Figure~\ref{fig:toy} compares H5, Tool, Web, Questionnaire, and Game browser-native artifacts under the same generation and evaluation process. Game tasks reach a 77.9\% usable rate, indicating that they are sufficiently buildable and evaluable; their excellent rate is only 21.5\%, indicating that minimal usability and complete requirement satisfaction remain distinct quality levels.

Based on this testbed, we propose WebGameBench, a requirement-to-application benchmark for coding agents. WebGameBench contains 111 browser-native game tasks spanning seven gameplay families. Each task is also assigned a specification-level difficulty label \(D\in\{D1,D2,D3,D4\}\), where larger \(D\)-levels indicate higher expected difficulty in generating a correct delivered application from the requirement. Each task uses a frozen Structured WebGame Specification as the sole task contract, specifying gameplay goals, interaction rules, state boundaries, deployment constraints, and requirement-derived functional points. The evaluation pipeline connects application synthesis, local deployment, and interactive runtime evaluation: the agent implements the application under a unified generation protocol; under the same generation and deployment protocol, the delivered application is exposed as a browser-accessible URL; and a runtime evaluator verifies the delivered application through real browser interaction, producing one of three labels, \excellent{}, \usablelabel{}, or \unusable{}, together with structured evidence.

This design evaluates coding agents' application-artifact delivery capability in the browser-native game setting, rather than only source-level correctness, build success, or page loading. Results on 12 coding agents and 14 evaluation configurations show interpretable model separation: the strongest configuration reaches a 76.9\% usable rate but only a 20.2\% excellent rate, indicating that crossing the usability threshold is not equivalent to satisfying the full requirement. Difficulty stratification also shows a marked drop in usable rate on high-risk tasks. A human-reviewed subset further shows that the runtime evaluator is reasonably aligned with human gameplay judgments under the Usable-rate criterion, while exact three-way label agreement remains substantially harder.

This paper makes three contributions. First, we formalize requirement-to-application evaluation and build WebGameBench, a closed-loop benchmark that connects specification-based generation, local deployment, and browser-interaction-based runtime evaluation for browser-native games. Second, we evaluate representative open and closed coding agents under a unified protocol and show that WebGameBench produces clear and interpretable model separation. Third, we validate the runtime evaluator against a human-reviewed subset, showing reasonable alignment with human gameplay judgments under the Usable-rate criterion while highlighting the difficulty of exact three-way quality labeling.

\section{Related Work}
\label{sec:related}

\paragraph{Code and software-engineering benchmarks.}
Function- and program-level benchmarks established executable correctness as the core signal for code-generation evaluation. HumanEval \citep{HumanEval-arXiv21}, MBPP \citep{MBPP-arXiv21}, APPS \citep{APPS-arXiv21}, and LiveCodeBench \citep{LiveCodeBench-ICLR25} cover function synthesis, basic program synthesis, competitive-programming tasks, and freshness / contamination-aware code evaluation. Repository-level and long-horizon engineering benchmarks such as SWE-bench \citep{SWE-bench-ICLR24}, SWT-Bench \citep{SWT-Bench-NeurIPS24}, Terminal-Bench \citep{Terminal-Bench-arXiv26}, SUPER \citep{SUPER-EMNLP24}, PaperBench \citep{PaperBench-arXiv25}, and MLE-bench \citep{MLE-bench-arXiv24} broaden evaluation to issue resolution, test generation, terminal work, research reproduction, and machine-learning engineering.

\paragraph{Web artifact generation and computer-use benchmarks.}
Web-generation benchmarks evaluate whether models can produce websites, front ends, or web applications from natural language, screenshots, sketches, or interaction prototypes \citep{WebGen-Bench-arXiv25,WebCoderBench-arXiv26,E2EDev-arXiv25,Design2Code-NAACL25,Sketch2Code-NAACL25,Interaction2Code-ASE25,WebUIBench-ACLFindings25}. Web and computer-use benchmarks instead evaluate agents operating existing browser, desktop, or mobile environments, including MIND2WEB \citep{MIND2WEB-NeurIPS23DB}, WebArena \citep{WebArena-ICLR24}, VisualWebArena \citep{VisualWebArena-ACL24}, WebVoyager \citep{WebVoyager-ACL24}, WorkArena \citep{WorkArena-arXiv24}, BrowserGym \citep{BrowserGym-arXiv24}, OSWorld \citep{OSWorld-NeurIPS24DB}, and AndroidWorld \citep{AndroidWorld-arXiv24}.

\paragraph{Game-interaction benchmarks.}
Games are widely used for evaluating planning, exploration, spatial reasoning, and long-horizon decision making in existing environments, including text games, general video-game frameworks, Minecraft-like settings, and recent LLM/VLM game suites \citep{TextWorld-arXiv18,Hausknecht2019InteractiveFG,GVGAI-TG19,MineRL-NeurIPS19,GameBench-arXiv24,BALROG-ICLR25,lmgame-Bench-arXiv25,TALES-arXiv25}. These benchmarks usually evaluate the agent as a player.

\section{Method}
\label{sec:method}

WebGameBench instantiates \emph{requirement-to-application evaluation} for browser-native games. Given a frozen structured requirement \(\sigma\), a coding agent \(A\) makes one generation attempt in a unified environment and produces a source artifact that is deployed as a browser-native application \(x=A(\sigma)\). Under the same generation and deployment protocol, the delivered application is exposed as a browser-accessible URL \(u\) for evaluation. A runtime evaluator \(E\) then judges, based on \((u,\sigma)\), whether the deployed application satisfies the requirement through browser interaction, returning a three-way quality label \(y\in\{\excellent{},\usablelabel{},\unusable{}\}\) together with structured evidence. The evaluated object is therefore the observable runtime behavior of the delivered application, rather than an intermediate reasoning trace, isolated code fragment, repository diff, or action sequence in an existing environment.

\begin{figure}[t]
  \centering
  \includegraphics[width=\linewidth]{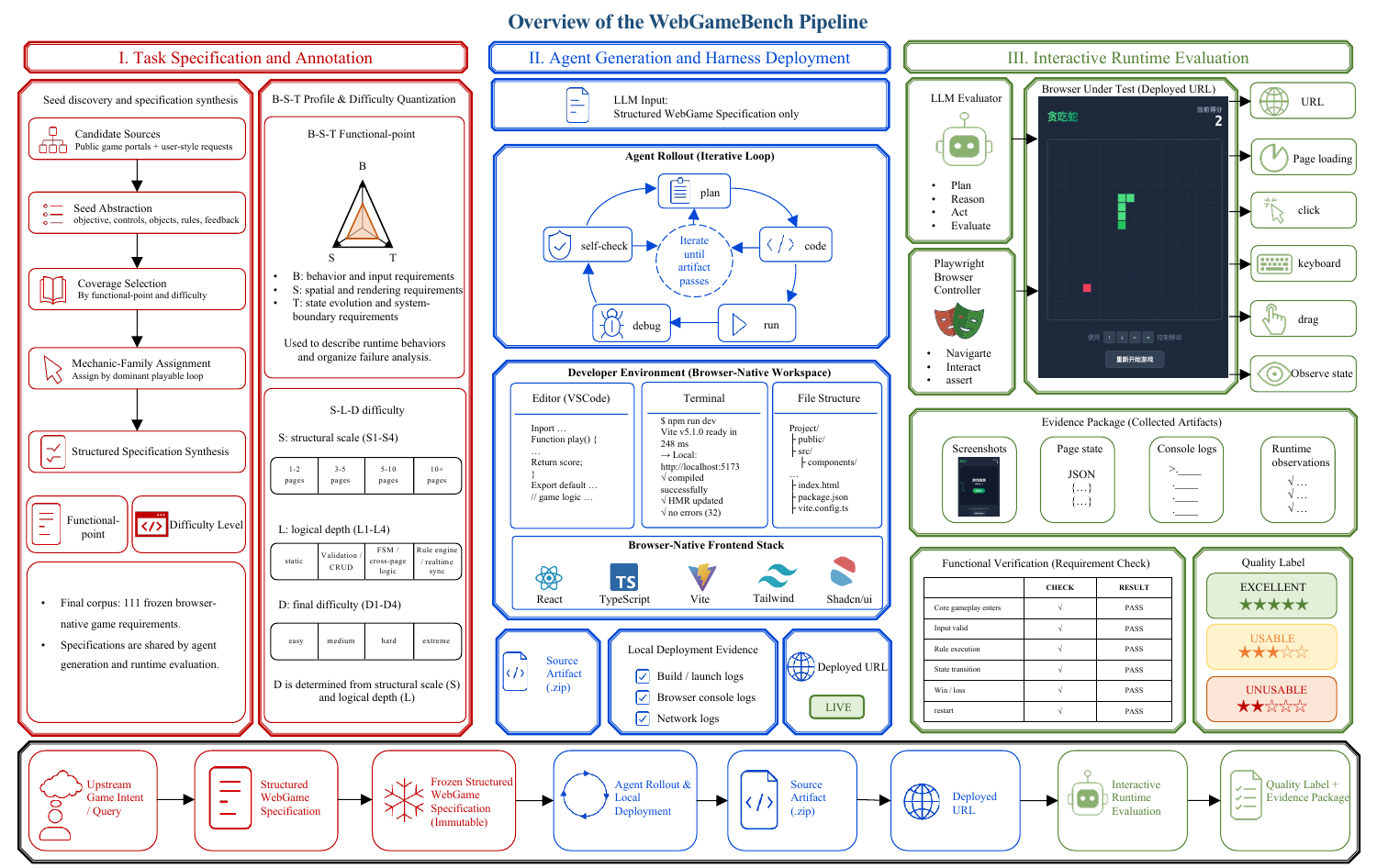}
  \caption{Overview of the WebGameBench pipeline. Each task is defined by a frozen Structured WebGame Specification; an agent produces a browser-native source artifact; the delivered application is exposed as a browser-accessible URL; and the runtime evaluator judges the delivered game against the same specification through real browser interaction.}
  \label{fig:pipeline}
\end{figure}

\subsection{Evaluation Setting and Protocol}
\label{sec:evaluation-framework}

Figure~\ref{fig:pipeline} summarizes the pipeline. The evaluation setting separates three roles that are often conflated in coding-agent benchmarks: task definition, application delivery, and runtime judgment. The Structured WebGame Specification fixes the requirement to be implemented; the coding agent attempts to deliver a browser-native application under a unified generation and deployment protocol; and the runtime evaluator judges the exposed URL against the same frozen requirement. This separation keeps the benchmark outcome tied to observable runtime behavior rather than to code appearance, build success alone, or evaluator-specific reinterpretation of the task.

\paragraph{Task specification and annotation.}
Each benchmark instance uses a frozen \emph{Structured WebGame Specification} as its formal task input, denoted by \(\sigma\). The specification follows a product-requirements-style organization while constraining the scope to browser-native game applications: it states the game objective, page or scene flow, input controls, game objects, rules and state transitions, visible feedback, deployment constraints, and observable acceptance criteria. Unlike the raw user request, \(\sigma\) is the shared fixed requirement representation used in evaluation: the coding agent relies on it to produce the delivered application, and the runtime evaluator judges the application against the same specification. During dataset construction, we also derive functional-point annotations and specification-level difficulty labels from \(\sigma\); both are defined in Section~\ref{sec:construction}.

\paragraph{Agent generation and deployment protocol.}
For each task and coding agent, WebGameBench runs one independent generation attempt in a standardized browser-native development environment. The agent receives the frozen specification \(\sigma\) under the same workspace template, tool interface, execution budget, and deployment protocol. The attempt produces a browser-native source artifact in a project workspace and exposes the delivered application as a browser-accessible URL \(u\). The generation record retains the source path, deployed URL, generation metadata, and trajectory log, so each delivered artifact can be traced back to the specification and agent attempt that produced it.

\paragraph{Interactive runtime evaluation.}
\label{sec:runtime}
Interactive runtime evaluation takes the browser-accessible URL \(u\) and the same frozen specification \(\sigma\) as input, and asks whether the delivered application satisfies the requirement in a real browser runtime. In our implementation, the runtime evaluator is a Codex-based agent that controls Chrome and executes browser interactions through Playwright. For the main results, we use Codex CLI 0.115.0 with the `gpt-5.4-agent' backend and XHigh reasoning effort; the human-review analysis additionally compares Medium, High, and XHigh reasoning settings. Each artifact receives one evaluation rollout with a two-hour wall-clock timeout and no additional fixed browser-action step cap. The evaluator plans the verification order from \(\sigma\), opens \(u\), performs user-level actions, reads page state, records runtime evidence, and produces a structured evaluation report. The evaluator is not a free-form subjective judge: each judgment must be grounded in the specification, real browser interaction, and runtime evidence.

The evaluation first determines whether the application is accessible and can enter an interactive state corresponding to the specification. It then verifies observable requirements around the main path, input response, spatial mapping, rule execution, state transition, score or resource change, terminal behavior, and restart behavior. Each judgment must be supported by user-visible page behavior, browser interaction outcomes, or stable runtime state; source code, build logs, and internal state are diagnostic evidence only, not substitutes for observable behavior. For long-horizon, randomized, spatial, or multi-session requirements, the evaluator may construct candidate preconditions or independent sessions, but the final judgment must still rest on interaction outcomes in the real browser. Finally, the runtime evaluator maps observed behavior to three quality labels: \excellent{}, \usablelabel{}, and \unusable{}, corresponding respectively to complete and stable satisfaction of main requirements, a playable main path with non-core defects, and failure to complete the core playable loop or to access the delivered application. These three labels form the label-valued runtime reward used by \benchmark{}; usable rate and excellent rate are aggregate statistics derived from this reward.

\subsection{Dataset Construction}
\label{sec:construction}

\benchmark{} constructs open but bounded browser-native game requirements. It does not collect existing web games as targets to copy, and it does not ask agents to reproduce a specific website, asset pack, or source implementation. The construction goal is a corpus of requirement-to-application tasks whose success and failure can be observed through browser runtime interaction.

\paragraph{Seed discovery and specification synthesis.}
Candidate game intents are derived from public browser-game ecosystems and real user requests. Public portals such as 4399, Poki, and CrazyGames are used only as taxonomy inspiration for gameplay families, interaction patterns, and session structures; their assets, layouts, level designs, text, and implementations are not copied. We normalize seed coverage into seven mechanic families: Puzzle/Matching, Action/Arcade/Platform, Shooting/Combat, Racing/Sports/Skill, Strategy/Simulation/Management, Casual/Reaction/Idle, and Board/Card/Local Multiplayer.

Each seed preserves the player objective, control scheme, core game objects, central rules, and observable feedback while removing branding, commercial IP, advertising, paid resources, external leaderboards, accounts, and remote services. Seeds and necessary upstream user material are then synthesized into frozen \specname{} documents. Each specification states the minimum playable loop: how the game starts, what inputs the player can execute, how objects respond, how score or resources change, which rules trigger state transitions, when the game terminates, and how the player restarts or continues.

\paragraph{Functional-point annotation.}
After specification normalization, we annotate each frozen Structured WebGame Specification with functional points by decomposing its observable requirements along three requirement-decomposition axes: behavior, spatial, and temporal/state. Behavior (\(B\)) covers player entry, input actions, and core interaction; spatial (\(S\)) covers position representation, coordinate mapping, boards, canvas regions, hit zones, and collision relations; and temporal/state (\(T\)) covers game entities, phases, random or hidden information, resource changes, terminal conditions, state transitions, and temporal dependencies. We organize these annotations in a four-tier B-S-T functional-point schema: Tier 1 contains the three axes \(B\), \(S\), and \(T\); Tier 2 defines stable subdimensions under each axis; Tier 3 defines controlled tags or enumerated values; and Tier 4 defines task-specific atomic functional points. An atomic functional point is an independently observable behavior or structural assertion, such as game entry, input response, spatial mapping, rule execution, state transition, score or resource update, terminal triggering, or restart flow. Functional-point annotation is used only as metadata for benchmark construction, coverage analysis, and experimental stratification; the complete schema, annotation rules, and corpus statistics are given in Appendix~\ref{app:functional-points}.

\paragraph{Difficulty labels.}
Difficulty labels characterize the expected implementation risk of a frozen Structured WebGame Specification before model evaluation. We assign this specification-level difficulty from two requirement-level axes: structural scale \(S_{\mathrm{struct}}\) and logical depth \(L_{\mathrm{logic}}\). The notation \(S_{\mathrm{struct}}\) distinguishes this difficulty axis from the spatial \(S\) in B-S-T annotation. \(S_{\mathrm{struct}}\) measures the breadth of pages, modules, navigation hierarchy, and user-role-specific interface scope, while \(L_{\mathrm{logic}}\) measures gameplay-rule depth, state-machine complexity, acceptance criteria, exception branches, persistence, AI behavior, and multi-session or synchronization constraints. The two axes are first assigned independently and then mapped by a fixed \(4\times4\) lookup table to \(D\in\{D1,D2,D3,D4\}\), where higher \(D\)-levels indicate higher expected implementation risk. The \(D\)-level is used for corpus profiling and difficulty-stratified analysis, not as a function of model outcomes or functional-point counts. The full annotation rubric, lookup rule, and stability analysis are given in Appendix~\ref{app:difficulty}.

\paragraph{Feasibility filtering and coverage selection.}
After specification synthesis and task annotation, we filter candidates into the final corpus. Filtering first screens for browser-native feasibility: the core playable loop must be self-contained, and success or failure must be observable through browser runtime evidence. We then apply coverage-aware selection over three complementary dimensions: seven gameplay families for mechanic coverage, B-S-T functional-point annotation for observable requirement coverage, and \(D\)-level labels for specification-level difficulty coverage. This process is not intended to produce a proportional sample of online games, but to form a requirement-to-application benchmark corpus that is bounded, evaluable, and diverse in gameplay mechanisms, observable functional points, and expected implementation risk.

\begin{figure}[t]
  \centering
  \includegraphics[width=\linewidth]{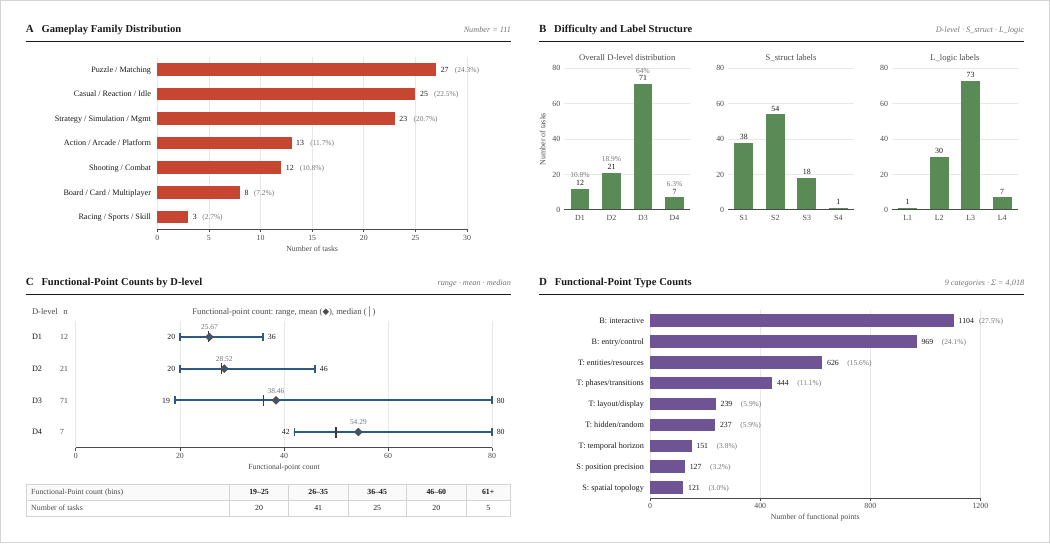}
  \caption{\benchmark{} corpus profile over 111 browser-native game requirements, including gameplay families, \(D\)-level labels, \(S_{\mathrm{struct}}\)/\(L_{\mathrm{logic}}\) axes, and B-S-T functional-point annotations.}
  \label{fig:corpus}
\end{figure}

\paragraph{Corpus profile.}
Figure~\ref{fig:corpus} summarizes the retained corpus. The seven gameplay families are coverage strata rather than proportional samples of online games. The corpus is structurally compact but logically dense: \(82.9\%\) of tasks have structural-scale S1 or S2 labels, while \(72.1\%\) have logical-depth L3 or L4 labels. This distribution places expected implementation risk more on gameplay rules, state transitions, scoring, terminal conditions, and multi-session behavior than on page count. The mean atomic functional-point count rises from 25.67 for D1 to 54.29 for D4, serving as a construction-time sanity check; B-S-T functional-point annotation is used for decomposition and profiling, not for assigning \(D\)-level.

\section{Experiments}
\label{sec:experiments}

\subsection{Experimental Setup}

All experiments use the same set of frozen \specname{} documents. For each evaluated coding agent, \benchmark{} runs the unified generation and deployment protocol defined in Section~\ref{sec:evaluation-framework}: the agent receives a frozen specification, produces a browser-native source artifact in a standardized workspace, and the artifact is built, served, and exposed as a browser-accessible URL under the same deployment protocol. During generation, all agents receive only the frozen specification and the common workspace template. The delivered application is then evaluated by the interactive runtime evaluator defined in Section~\ref{sec:runtime}. API batching concurrency, timeouts, and trace statistics are reported in Appendix~\ref{app:compute}.

We report 14 evaluation configurations covering 12 representative coding agents. The count is larger than the number of agents because DeepSeek-V4 Pro and DeepSeek-V4 Flash are each evaluated under thinking and non-thinking inference settings; these rows are configuration variants within the same model family rather than additional model families. The evaluated configurations cover Anthropic Claude Opus/Sonnet models \citep{AnthropicClaudeOpus47,AnthropicClaudeOpus46,AnthropicClaudeSonnet45}, OpenAI GPT-5.5 \citep{OpenAI-GPT55}, Google Gemini 3.1 Pro \citep{GoogleGemini31Pro}, DeepSeek-V4 \citep{deepseekai2026deepseekv4}, Kimi K2.6 \citep{KimiK26TechBlog}, Kimi K2.5 \citep{KimiK25-arXiv26}, GLM-5/5.1 \citep{GLM5-arXiv26,ZAI-GLM51}, and Tencent Hy3 \citep{TencentHy3Preview}. All configurations use the same input specifications, generation protocol, and runtime evaluation prompt.

The evaluator returns \excellent{}, \usablelabel{}, or \unusable{}. We report coverage, \textbf{Excellent rate} \(R_{\mathrm{exc}}\), and \textbf{Usable rate} \(R_{\mathrm{use}}\). Coverage is the fraction of evaluation attempts with a valid returned label. Artifact-attributable installation, build, serving, or access failures are valid \unusable{} labels; invalid labels are limited to evaluator or deployment infrastructure failures, corrupted evidence packages, or unresolved evaluator timeouts that prevent reliable scoring. \(R_{\mathrm{exc}}\) is the fraction of valid scored samples labeled \excellent{}. \(R_{\mathrm{use}}\) is the fraction of valid scored samples labeled \excellent{} or \usablelabel{}, i.e., whether the artifact crosses the minimum playable-delivery threshold.

\subsection{Model Capability Differentiation}

\begin{table}[t]
\centering
\scriptsize
\caption{Main results on \benchmark{}. Excellent rate and Usable rate use valid scored samples as the denominator; D1--D4 columns report usable/valid counts at each \(D\)-level.}
\label{tab:main}
\resizebox{\linewidth}{!}{%
\begin{tabular}{lrrrrrrr}
\toprule
Configuration & Coverage & Excellent rate & Usable rate & \multicolumn{4}{c}{Usable / valid by \(D\)-level} \\
\cmidrule(lr){5-8}
 & & & & D1 & D2 & D3 & D4 \\
\midrule
opus-4-7 \citep{AnthropicClaudeOpus47} & 93.7\% & 20.2\% & 76.9\% & 11/12 & 17/21 & 49/65 & 3/6 \\
opus-4-6 \citep{AnthropicClaudeOpus46} & 90.1\% & 19.0\% & 73.0\% & 9/12 & 17/18 & 45/63 & 2/7 \\
gpt-5-5 \citep{OpenAI-GPT55} & 99.1\% & 16.4\% & 63.6\% & 8/12 & 17/20 & 43/71 & 2/7 \\
gemini-3.1-pro \citep{GoogleGemini31Pro} & 96.4\% & 15.9\% & 63.6\% & 10/12 & 15/20 & 42/68 & 1/7 \\
deepseek-v4-flash-thinking \citep{deepseekai2026deepseekv4} & 99.1\% & 13.6\% & 62.7\% & 11/12 & 17/21 & 40/70 & 1/7 \\
deepseek-v4-pro-thinking \citep{deepseekai2026deepseekv4} & 100.0\% & 12.6\% & 62.2\% & 8/12 & 18/21 & 43/71 & 0/7 \\
kimi-k2.6 \citep{KimiK26TechBlog} & 94.6\% & 16.2\% & 61.0\% & 8/12 & 15/21 & 41/65 & 0/7 \\
deepseek-v4-flash-nothinking \citep{deepseekai2026deepseekv4} & 97.3\% & 14.8\% & 59.3\% & 11/12 & 17/21 & 35/68 & 1/7 \\
GLM-5.1 \citep{ZAI-GLM51} & 97.3\% & 12.0\% & 52.8\% & 7/12 & 19/21 & 31/68 & 0/7 \\
GLM-5 \citep{GLM5-arXiv26} & 99.1\% & 10.9\% & 48.2\% & 8/12 & 14/21 & 31/70 & 0/7 \\
sonnet-4-5 \citep{AnthropicClaudeSonnet45} & 91.9\% & 7.8\% & 47.1\% & 7/12 & 15/21 & 25/63 & 1/6 \\
deepseek-v4-pro-nothinking \citep{deepseekai2026deepseekv4} & 100.0\% & 16.2\% & 44.1\% & 9/12 & 13/21 & 26/71 & 1/7 \\
hy3-xhigh \citep{TencentHy3Preview} & 95.5\% & 6.6\% & 41.5\% & 9/12 & 14/21 & 21/67 & 0/6 \\
kimi-k2.5 \citep{KimiK25-arXiv26} & 96.4\% & 8.4\% & 38.3\% & 7/11 & 12/21 & 22/68 & 0/7 \\
\bottomrule
\end{tabular}
}
\end{table}

Table~\ref{tab:main} shows that \benchmark{} produces clear capability separation among current coding agents. The strongest two configurations, `opus-4-7' and `opus-4-6', reach \(R_{\mathrm{use}}\) values of 76.9\% and 73.0\%, while lower-scoring configurations fall in the 38.3--52.8\% range. Middle configurations form a continuous band around 60\%, indicating that the benchmark does not collapse all configurations into the same success or failure region, but instead provides interpretable model differences.

\textbf{Usable rate} \(R_{\mathrm{use}}\) indicates whether an artifact is labeled \excellent{} or \usablelabel{}, but it is not a high-quality completion metric. Even the best configuration still leaves roughly one quarter of tasks labeled \unusable{}. More importantly, \textbf{Excellent rate} \(R_{\mathrm{exc}}\) is much lower: the highest \(R_{\mathrm{exc}}\) is only 20.2\%, and most configurations are below 17\%. This gap shows that crossing the minimum playable-delivery threshold is substantially easier than satisfying the full set of complex runtime requirements.

The D1--D4 columns further show that the construction-time \(D\)-level is directionally aligned with runtime usable rate. Aggregating the usable/valid counts in Table~\ref{tab:main} by \(D\)-level gives pooled usable rates of \(123/167=73.7\%\), \(220/289=76.1\%\), \(494/948=52.1\%\), and \(12/95=12.6\%\) for D1, D2, D3, and D4, respectively. Thus D1/D2 tasks remain near a 75\% usable rate, D3 drops to about 52\%, and D4 reaches only 12.6\%. We therefore interpret \(D\)-level as a coarse risk stratification useful for explaining shared failures on D3/D4 tasks, rather than as a strict total order over individual task difficulty.

\subsection{Agreement with Human Review}

We evaluate whether the runtime evaluator agrees with independent human gameplay review on a 43-artifact human-review set covering Medium, High, and XHigh reasoning settings. Human review was performed by five annotators, with two independent human tests per artifact. Human labels follow a playable-lifecycle criterion: a game is usable if the player can enter or start it, exercise the core input, receive rule feedback, observe state or score updates, and reach a win/loss or ending state; failures to start or end, mid-game deadlocks, broken core input, and incorrect win/loss logic are marked \unusable{}. Two senior experts then inspected disagreements between human labels and automatic evaluator outputs, focusing on whether they arise at the \(R_{\mathrm{use}}\) boundary between \excellent{}/\usablelabel{} and \unusable{}, or at the \(R_{\mathrm{exc}}\) boundary between \excellent{} and \usablelabel{}.

\begin{figure}[t]
  \centering
  \includegraphics[width=\linewidth]{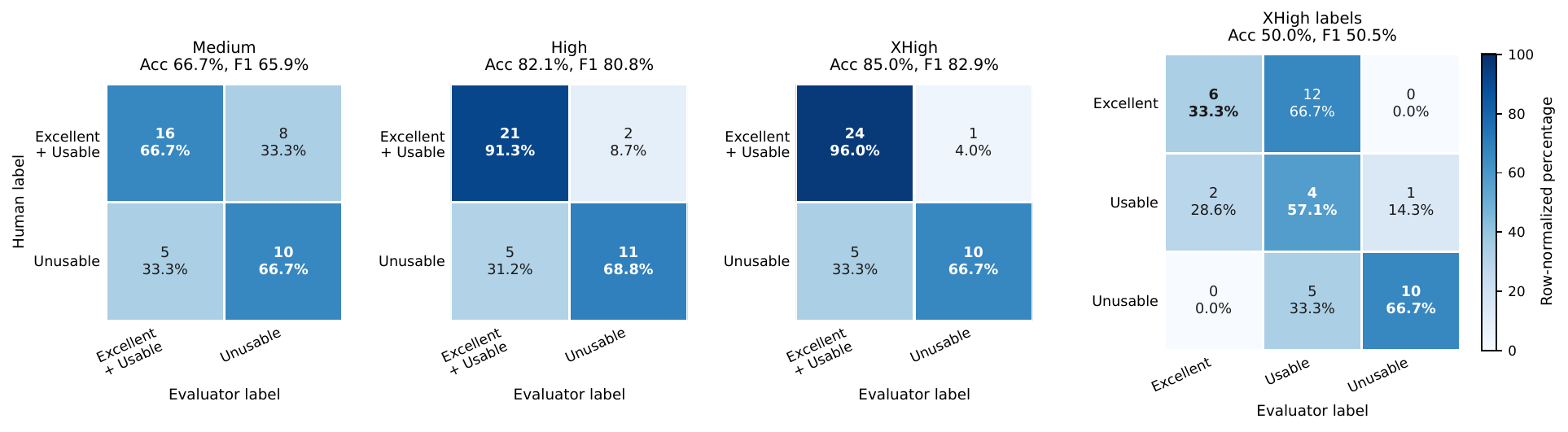}
  \caption{Agreement between the runtime evaluator and human-review labels. The first three heatmaps use the \(R_{\mathrm{use}}\) label, merging \excellent{} and \usablelabel{} into \emph{Excellent + Usable}; the fourth keeps the three labels under XHigh to inspect the \(R_{\mathrm{exc}}\) boundary. Colors are row-normalized by the human label, and cell text reports counts and row percentages.}
  \label{fig:human}
\end{figure}

Figure~\ref{fig:human} reports agreement for the \(R_{\mathrm{use}}\) label and for exact three-way labels under XHigh. Agreement on \(R_{\mathrm{use}}\) improves with evaluator reasoning strength: Medium reaches 66.7\% accuracy and 65.9\% macro-F1, High reaches 82.1\% and 80.8\%, and XHigh reaches 85.0\% and 82.9\%, supporting the evaluator as a scalable signal for Usable-rate statistics. Exact three-way agreement remains lower under XHigh, at 50.0\% accuracy and 50.5\% macro-F1, mainly because the \(R_{\mathrm{exc}}\) boundary is harder: 12 samples labeled \excellent{} by humans are downgraded to \usablelabel{} by the evaluator. We therefore use automatic evaluation for aggregate Usable-rate statistics and runtime failure diagnosis, while treating Excellent-rate analysis as a stricter quality signal that still benefits from human review.

\subsection{Runtime Case Studies}
\label{sec:runtime_case_studies}

\begin{figure}[H]
\centering
\makebox[\linewidth][c]{\includegraphics[width=1.12\linewidth]{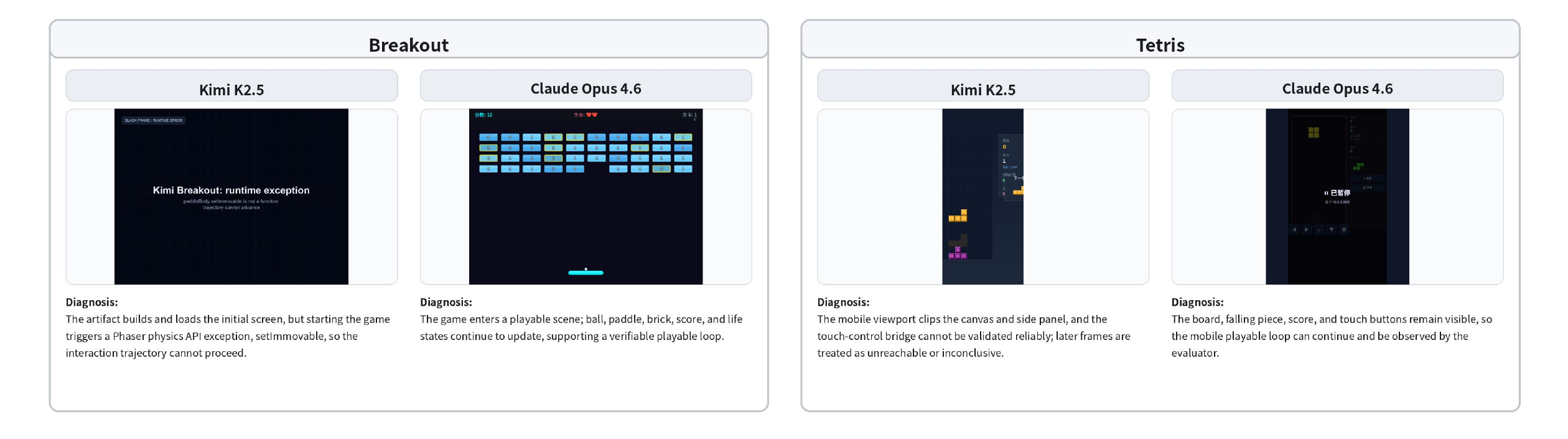}}
\caption{Two representative runtime case studies in the main text. Each task compares Kimi K2.5 and Claude Opus 4.6, with evaluator diagnosis shown below the corresponding screenshots.}
\label{fig:case-study}
\end{figure}

To connect aggregate labels to runtime evidence, we analyze paired traces from Kimi K2.5 and Claude Opus 4.6 on six games; Figure~\ref{fig:case-study} shows the two contrast cases most directly tied to our main claim, with the remaining cases in Appendix~\ref{app:supplementary_runtime_cases}.

Figure~\ref{fig:case-study} illustrates why buildability and loadability are not application usability. In Breakout, the Kimi artifact loads but fails after start because a Phaser physics exception breaks the control loop, while the Opus artifact preserves ball, paddle, brick, score, and life updates. In Tetris, Kimi's mobile layout and touch bridge prevent stable playable evidence, while Opus keeps the board, falling pieces, score, and controls observable. These cases show that runtime evaluation diagnoses interaction, state-transition, and rule-execution failures that build logs or screenshots can miss.

\section{Limitations and Discussion}
\label{sec:limitations}

\benchmark{} has several limitations. The runtime evaluator supports scalable Usable-rate estimation and failure diagnosis, but it is not a substitute for human review: agreement on the Usable-rate criterion is reasonable, while exact three-way quality labeling remains limited. The benchmark is instantiated with browser-native games, so its results should be read as evidence about browser-accessible interactive artifact delivery rather than arbitrary software-engineering competence. The frozen specifications are produced through fixed templates and filtering, and candidate precondition construction cannot exhaust long-horizon, randomized, or multi-session behavior. We therefore report aggregate scores together with structured evidence and human-review analysis, and position \benchmark{} as a diagnostic complement to human review rather than a standalone certification of application quality.

\section{Conclusion}
\label{sec:conclusion}

We introduced \benchmark{}, a requirement-to-application benchmark for coding agents based on browser-native games. Each task provides a frozen \specname{}; the agent generates a source artifact; the artifact is built, served, and exposed as a browser-accessible URL under the same deployment protocol; and a runtime evaluator verifies input handling, spatial mapping, rule execution, state transitions, terminal conditions, and restart behavior through real browser interaction. Experiments on 111 tasks, 12 coding agents, and 14 evaluation configurations show that \benchmark{} differentiates current models' delivered-application capability in this setting: the best configuration reaches a 76.9\% usable rate, but only a 20.2\% excellent rate, showing that crossing the minimum playable-delivery threshold is not equivalent to complete requirement satisfaction. Further \(D\)-level stratification, human review, and runtime case studies show that \benchmark{} supports aggregate scoring, Usable-rate human-alignment analysis, and runtime failure diagnosis. Overall, browser-native games provide a compact and behavior-dense testbed for evaluating whether coding agents can deliver runnable, interactive, and diagnosable software artifacts from structured requirements.

\bibliographystyle{unsrtnat}
\bibliography{references}

\appendix
\section{Specification Construction Details}
\label{app:construction}

This appendix records construction details that are intentionally compressed in the main text. The main text defines the Structured WebGame Specification as the shared fixed requirement representation used in evaluation. Here we explain how upstream materials are abstracted into that task input. In the engineering pipeline, this frozen requirement document was sometimes called a PRD. In the paper, we consistently call it a Structured WebGame Specification: a task contract produced from an original user request or seed through a fixed template, and shared by both agent generation and runtime evaluation.

\subsection{Seed Abstraction}

WebGameBench uses public browser-game portals and user-style requests only as inspiration for gameplay families and interaction patterns. A candidate page or upstream user request is first reduced to a task seed: it preserves the playable concept while removing brand names, site-specific copy, advertisements, accounts, leaderboards, concrete art assets, audio, level files, implementation details, and external service dependencies. The seed is not the benchmark input. It is an intermediate construction artifact used to synthesize a self-contained Structured WebGame Specification.

\subsection{Specification Template}

Each Structured WebGame Specification follows a fixed product-requirements-style template:
\begin{itemize}
    \item game overview: name, type, one-sentence description, and player objective;
    \item page and flow: initial page, game page, ending or settlement page, and restart;
    \item input controls: primary input, auxiliary input, and invalid-input feedback;
    \item game objects: player objects, non-player objects, board or scene, and random objects;
    \item rules and state changes: initial state, per-frame or per-turn changes, valid-action effects, score or resource changes, difficulty changes, failure state, and victory or level-clear state;
    \item visual feedback: layout, state display, operation feedback, success feedback, and failure feedback;
    \item technical and deployment constraints: running mode, external dependencies, storage, devices and browsers, and forbidden items;
    \item evaluable functional points: loading, start or restart, input response, core rules, observable state changes, score/resource/progress updates, terminal state, optional victory or level-clear state, and no dependence on private assets or remote services.
\end{itemize}

\subsection{Specification Synthesis Prompt}

The specification synthesis prompt asks a model to rewrite a short user request or task seed into a complete, implementable, and evaluable Structured WebGame Specification, rather than implementation code. The prompt uses a fixed section skeleton: application overview, users and usage scenarios when applicable, page structure and core functions, business rules and logic, exceptions and boundary cases, acceptance criteria, and out-of-scope functionality. For game tasks, the prompt further requires background and goals to serve gameplay description, core functions to center on the minimum playable loop, business rules to express gameplay rules and win/loss or ending conditions, and page or level structure to remain reasonably simple while preserving key exceptions and boundary cases.

The prompt enforces four constraints. First, it preserves the game identity, core gameplay, and all details already given in the user input. Second, it follows an MVP boundary: it only adds capabilities that are necessary for the core objective, and does not add accounts, leaderboards, payments, remote services, future plans, or other unrelated extensions. Third, it does not specify a concrete technology stack or implementation path. Fourth, after entering the rewrite stage, it does not ask clarification questions and directly outputs a frozen requirement document in Markdown. The generated document is the Structured WebGame Specification used in this paper. Functional-point annotation, difficulty annotation, agent generation, and runtime evaluation all use this frozen specification as input.

\section{Functional-Point Annotation}
\label{app:functional-points}

The main text uses functional points as derived annotations for observable requirement decomposition and corpus profiling. This appendix gives the B-S-T functional-point schema, the atomic functional-point construction rules, and corpus-level statistics.

\subsection{B-S-T Functional-Point Schema}

Functional-point annotation starts from three observable requirement-decomposition axes and organizes them in a four-tier schema. In Tier 1, \(B\) denotes behavior, describing player entry, input actions, and core interaction; \(S\) denotes spatial structure, describing position representation, coordinate mapping, boards, canvas regions, hit zones, and collision relations; and \(T\) denotes temporal/state behavior, describing game entities, phases, random or hidden information, resource changes, terminal conditions, state transitions, and temporal dependencies. Tier 2 defines stable subdimensions under each Tier-1 axis; Tier 3 defines controlled tags or enumerated values under each subdimension; and Tier 4 is the lowest-level task-specific functional point, namely an independently observable atomic behavior or structural assertion.

\begin{center}
\small
\begin{tabular}{lp{0.25\linewidth}p{0.50\linewidth}}
\toprule
Tier & Meaning & Examples \\
\midrule
Tier 1 & Broad decomposition axis & \(B\), \(S\), \(T\) \\
Tier 2 & Stable subdimension & \(B\): entry/controls and interaction; \(S\): topology and position precision; \(T\): layout/display, entities/resources, phases/transitions, hidden/random information, and temporal horizon \\
Tier 3 & Controlled tag or value & entry, input response, spatial mapping, collision, state transition, terminal condition \\
Tier 4 & Atomic functional point & ``Pressing the left arrow moves the player left''; ``The game shows a restart control after failure'' \\
\bottomrule
\end{tabular}
\end{center}

\subsection{Atomic Functional-Point Rules}

An atomic functional point is the counting unit used in corpus profiling. Each item must be atomic, independently observable, and grounded in the frozen Structured WebGame Specification. If one requirement contains multiple observable behaviors, it is decomposed into multiple atomic functional points. For example, a four-direction movement requirement can be split into separate items for moving up, down, left, and right; a complete terminal flow can be split into failure detection, settlement display, and restart entry.

Each atomic functional point is associated with one axis/subdimension/tag path and bound to a concrete UI or interaction component. If the assertion describes overall topology or global state behavior, it is bound to the global game structure. Pure visual polish is not counted as an independent atomic functional point unless the specification explicitly makes it an observable success condition. This keeps functional points focused on behavior, spatial mapping, state change, and user-visible feedback rather than implementation style.

\subsection{Corpus Statistics}

Using B-S-T functional-point annotation, the final 111 tasks contain 19--80 atomic functional points per task, with mean 36.20 and median 33. B-S-T annotation is used only for decomposition and corpus profiling, not for assigning the difficulty label. The \(D\)-level is determined separately by the \(S_{\mathrm{struct}}\)-\(L_{\mathrm{logic}}\) lookup rule in Appendix~\ref{app:difficulty}. When grouped by \(D\)-level, functional-point count increases on average, which serves as a corpus-level sanity check rather than a labeling rule.

\begin{center}
\small
\begin{tabular}{lrrrrr}
\toprule
\(D\)-level & N & Min & Max & Mean & Median \\
\midrule
D1 & 12 & 20 & 36 & 25.67 & 25.5 \\
D2 & 21 & 20 & 46 & 28.52 & 28 \\
D3 & 71 & 19 & 80 & 38.46 & 36 \\
D4 & 7  & 42 & 80 & 54.29 & 50 \\
\bottomrule
\end{tabular}
\end{center}

\section{Difficulty Annotation}
\label{app:difficulty}

The main text uses \(D\)-level as a coarse difficulty tag for experimental stratification and corpus profiling. This appendix gives the \(S_{\mathrm{struct}}\)-\(L_{\mathrm{logic}}\) rubric, lookup rule, and stability check used to determine \(D\)-level.

\subsection{Structural Scale and Logical Depth}

Difficulty annotation first assigns two axes independently from the frozen Structured WebGame Specification. \(S_{\mathrm{struct}}\) measures UI and navigation scale: S1 denotes 1--2 pages, flat structure, and no nesting; S2 denotes 3--5 pages, linear navigation, and standard component stacking; S3 denotes 5--10 pages with secondary nesting, global component extraction, or region partitioning, and also includes applications with more than 10 pages but shallow module hierarchy; S4 denotes more than 10 pages with deep module hierarchy or multiple independent role-specific portals. \(L_{\mathrm{logic}}\) measures rule and acceptance depth: L1 denotes static or nearly static presentation, no substantive state management, and fewer than 5 acceptance criteria; L2 denotes form validation, CRUD, persistence, or simple linear state transitions, with roughly 5--15 acceptance criteria; L3 denotes more than 15 acceptance criteria, complex state machines, conditional branches or rollback paths, or cross-module consistency constraints; L4 denotes complex algorithms, core rule engines, strong consistency, or multi-session real-time synchronization.

\subsection{D-level Lookup Rule}

After the two axes are assigned, \(D\)-level is obtained by a fixed lookup table:

\begin{center}
\small
\begin{tabular}{lcccc}
\toprule
\(S_{\mathrm{struct}}/L_{\mathrm{logic}}\) & L1 & L2 & L3 & L4 \\
\midrule
S1 & D1 & D1 & D3 & D3 \\
S2 & D1 & D2 & D3 & D3 \\
S3 & D2 & D2 & D3 & D4 \\
S4 & D2 & D3 & D4 & D4 \\
\bottomrule
\end{tabular}
\end{center}

The lookup rule makes \(D\)-level deterministic once \(S_{\mathrm{struct}}\) and \(L_{\mathrm{logic}}\) are determined, and intentionally makes logical depth dominant. Thus compact tasks with L3/L4 logic enter at least D3, while multi-page tasks with standard L2 logic can remain D2.

\subsection{Operational Annotation and Stability}

The annotation workflow first extracts evidence for page count, module hierarchy, and user-role-specific interface scope from the specification's page-structure and core-function sections, and assigns \(S_{\mathrm{struct}}\). It then extracts evidence for rule count, exception branches, state machines, and acceptance-criteria count from the rules, exceptions, boundary cases, and acceptance sections, and assigns \(L_{\mathrm{logic}}\). Finally, it strictly applies the lookup rule to output the \(D\)-level and a one-sentence implementation-risk summary. The final prompt explicitly covers all 16 \(S_{\mathrm{struct}}\)-\(L_{\mathrm{logic}}\) combinations and adds boundary rules to reduce ambiguity for S1+L2, S2+L3, S3+L2, and related cases. We check stability through repeated annotation and cross-prompt-version comparison. On the benchmark data, the final version reaches approximately 90\% single-request \(D\)-level agreement against majority labels from repeated runs.

\subsection{Difficulty Annotation Prompt}

The final difficulty annotation prompt is:
\begin{quote}
\small
You are a senior software architect responsible for quantitative complexity grading of a Structured WebGame Specification. Analyze the breadth of UI structure \(S\) and the depth of business logic \(L\), and determine the final difficulty \(D\). Larger numbers indicate higher implementation risk.

\textbf{\(S\) structural scale.} S1: 1--2 pages, flat structure, no nesting. S2: 3--5 pages, linear navigation, standard component stacking. S3: either 5--10 pages with secondary nesting, global component extraction, or region partitioning; or more than 10 pages with shallow module hierarchy. S4: more than 10 pages and either deep module hierarchy or multiple independent role-specific portals.

\textbf{\(L\) logical depth.} L1: static or nearly static presentation, no substantive state management, fewer than 5 acceptance criteria. L2: form validation, CRUD, persistence, or simple linear state transition, usually 5--15 acceptance criteria. L3: more than 15 acceptance criteria, or complex state machines, conditional branches, rollback paths, or cross-module consistency constraints. L4: complex algorithms, core rule engines, strong consistency, or multi-session real-time synchronization.

First determine \(S\) and \(L\) independently from the frozen specification, then obtain \(D\) strictly from the lookup rule in Appendix~\ref{app:difficulty}; do not infer it subjectively. Boundary rules: when \(L\geq L3\), \(D\) is at least D3; S1+L2 is D1; S1/S2+L3 is D3; S3/S4+L2 is D2; D4 appears only for S3+L4 or S4+L3/L4. Do not use model generation results, runtime evaluator results, or B-S-T atomic functional-point counts to determine \(D\). Output only three lines: \texttt{S: \{S1--S4\} | evidence}, \texttt{L: \{L1--L4\} | evidence}, and \texttt{D: \{D1--D4\} | lookup evidence}.
\end{quote}

\section{Runtime Evaluator Prompt and Report Schema}
\label{app:runtime_prompt}

The main text defines the runtime evaluation protocol. This appendix records the evaluator's operational prompt schema. Main experiments use Codex CLI 0.115.0 with the `gpt-5.4-agent' backend at XHigh reasoning effort, a two-hour wall-clock timeout per evaluation rollout, and no additional fixed browser-action step cap. The human-review agreement analysis reruns the same protocol under Medium, High, and XHigh reasoning settings.

\subsection{Evaluator Inputs}

Each evaluation receives:
\begin{itemize}
    \item a browser-accessible deployed URL;
    \item the frozen \specname{} for the same task;
    \item optional source bundle, build logs, serving logs, browser logs, screenshots, traces, and generation metadata for diagnosis.
\end{itemize}
The evaluator does not receive a reference implementation and does not provide feedback to the same agent attempt.

\subsection{Interaction Rules}

The evaluator prompt enforces the following order:
\begin{enumerate}
    \item open the deployed URL with Playwright and handle browser-access issues when needed;
    \item perform a short smoke interaction to check loadability, entry, and the main playable state;
    \item derive checks from the frozen specification and generic playable-loop requirements;
    \item verify checks through user-level actions before relying on source code or logs;
    \item record the pre-trigger state, trigger action, visible result, and relevant state or numeric deltas;
    \item mark each acceptance item as passed, failed, or unverified.
\end{enumerate}
Source code may explain a failure, but it cannot by itself prove that a user-visible behavior works.

\subsection{Candidate Preconditions}

For long-horizon, rare, or randomized states, code or runtime-state manipulation may be used to construct a candidate precondition. This operation does not itself verify the target function. The evaluator must first confirm that the candidate state is visible on the page or stably readable from runtime state, and then execute the final step through real user interaction. If the candidate state cannot be confirmed, the item is recorded as failed to construct, unstable to reproduce, or requiring review, rather than directly counted as a functional failure.

\subsection{Spatial and Multi-Session Rules}

Spatial checks must compare the intended operation location with the actual effect location. For boards, grids, canvas regions, drag slots, hit zones, placement cells, and connection points, a reaction is insufficient: the visible operation point and the effective point must agree. If visual interaction conflicts with DOM or state inspection, the evaluator prioritizes the result closer to normal user operation.

Multi-session checks are enabled only when required by the specification. The evaluator must construct independent sessions, record the visible state in each session, and check room creation, joining, start conditions, turn/state synchronization, settlement consistency, exit, refresh, and ownership behavior according to the requirement. A single endpoint cannot prove a multi-session mechanism.

\subsection{Report Schema}

Each report contains:
\begin{enumerate}
    \item final conclusion: row or sample id, deployed URL, source reference, quality label, and one-sentence summary;
    \item core functional checks: pass, fail, or unverified for the main playable loop;
    \item other issues: non-core failures and whether they affect the final label;
    \item acceptance results: per-requirement observations and evidence;
    \item unverified items: reason, attempted interactions, and whether review is needed.
\end{enumerate}

\section{Compute Resources and Trace Statistics}
\label{app:compute}

All generation and evaluation jobs are executed through API calls on a CPU cluster. We do not train models or perform GPU fine-tuning. Agent generation uses concurrency 6, and runtime evaluation uses concurrency 20; each concurrent job corresponds to one independent CPU-cluster execution unit. The per-sample timeout is 2 hours for both generation and evaluation. Because different coding-agent APIs have different service-side implementations, latency, and pricing, we do not report exact per-model dollar cost. Instead, we use turn counts from generation and evaluation traces as a compute-effort proxy.

Table~\ref{tab:trace_stats} reports available traces for 111 game tasks in the \benchmark{} benchmark data under selected model configurations. The table is used to confirm the data status and scale of API batch runs, not as full per-model cost accounting. Each row corresponds to one coding-agent configuration evaluated on the same benchmark tasks. The Opus 4.6 row corresponds to the `opus-4-6' generation-model configuration in the main experiment. Generation turns count assistant response turns in the generation trajectory; evaluation turns count assistant/agent messages in the runtime evaluator rollout. This table reports only sampled configurations used for data inspection and does not replace the full model comparison in Table~\ref{tab:main}. Game N is the number of benchmark game tasks for that configuration; Gen N and Eval N are the numbers of samples with parsed generation trajectories and evaluation rollouts; mean, med., P90, and max denote the mean, median, 90th percentile, and maximum turn counts per sample.

\begin{table}[h]
\centering
\scriptsize
\setlength{\tabcolsep}{3.5pt}
\caption{Generation/evaluation trace turn statistics for selected model configurations on \benchmark{} benchmark data. P90 denotes the 90th-percentile turn count.}
\label{tab:trace_stats}
\resizebox{\linewidth}{!}{%
\begin{tabular}{lrrrrrrrrrr}
\toprule
Model & Game N & Gen N & Gen mean & Gen med. & Gen P90 & Gen max & Eval N & Eval mean & Eval med. & Eval P90 / max \\
\midrule
Opus 4.6 & 111 & 111 & 20.3 & 19 & 36 & 47 & 100 & 34.8 & 34.0 & 49 / 75 \\
Sonnet 4.5 & 111 & 111 & 22.1 & 19 & 39 & 63 & 102 & 33.9 & 32.5 & 49 / 63 \\
Kimi K2.5 & 111 & 111 & 23.7 & 20 & 45 & 66 & 107 & 31.2 & 29.0 & 44 / 68 \\
\bottomrule
\end{tabular}
}
\end{table}

Overall, the three inspected configurations all have complete generation traces, with average generation turns around 20--24. Parsed evaluation rollouts cover 100--107 game tasks, with average evaluation turns around 31--35. These results indicate that the main experimental batch runs are dominated by bounded API generation and browser runtime evaluation, with interaction scale in the tens of turns, rather than by manual long-horizon debugging or open-ended repeated attempts.

\section{Supplementary Runtime Case Studies}
\label{app:supplementary_runtime_cases}

The main text keeps only the Breakout and Tetris contrast cases in Figure~\ref{fig:case-study}. This section supplements the remaining four tasks from case-study v3, covering boundary cases that are not expanded in the main text: entry failure with partially verifiable core rules, positive playable traces, dependency blockage, and missing evidence.

\begin{figure}[H]
\centering
\makebox[\linewidth][c]{\includegraphics[width=1.12\linewidth]{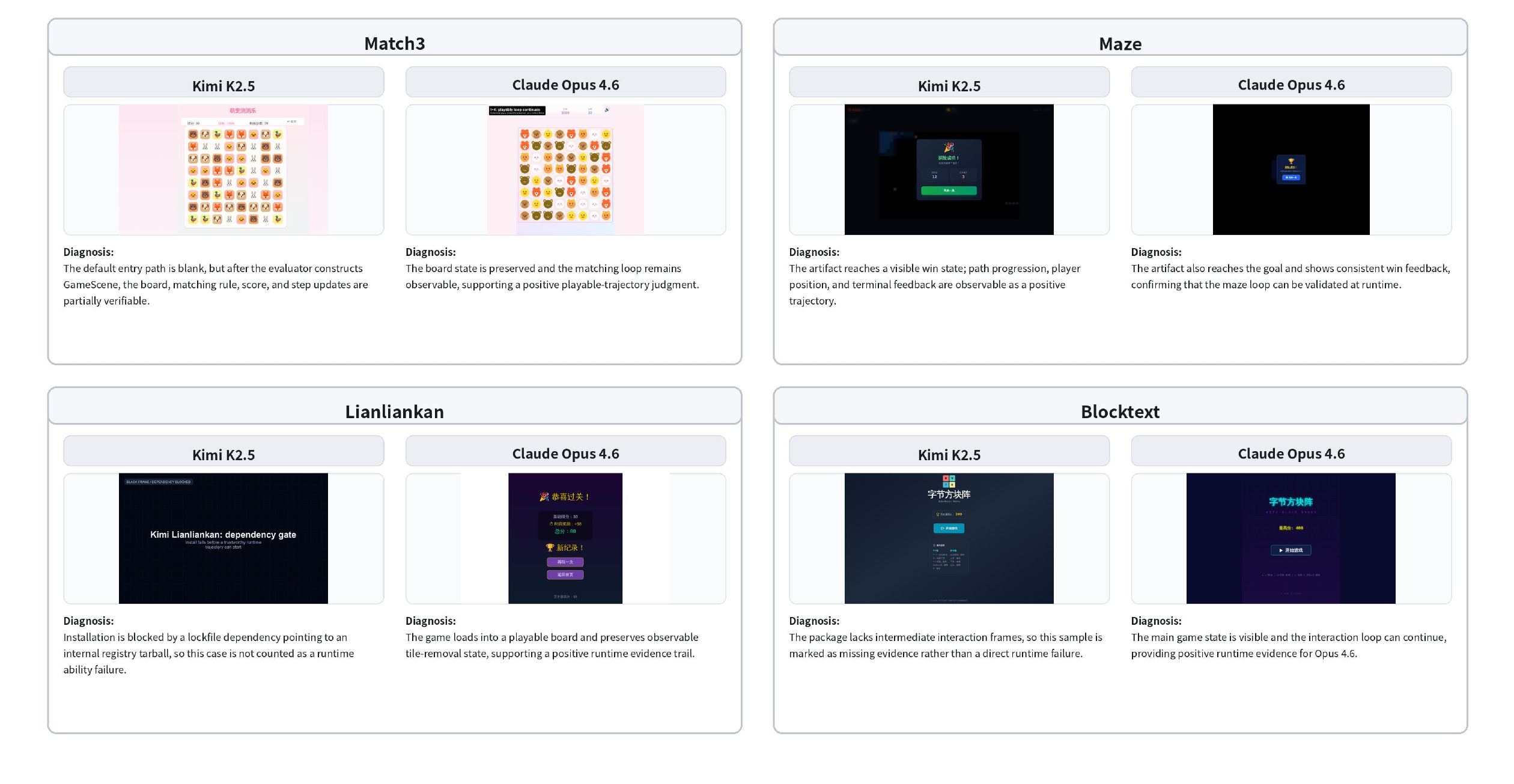}}
\caption{Supplementary runtime case studies for the four tasks omitted from the main text. Each task compares Kimi K2.5 and Claude Opus 4.6 with one representative screenshot and an evaluator diagnosis.}
\label{fig:case-study-appendix}
\end{figure}

\paragraph{Match3.} Kimi K2.5's default entry path is blank, but after the evaluator constructs entry into GameScene, the board, matching rules, score, and step updates can be partially verified. This sample should therefore not be treated as a pure blank-screen failure; it is better recorded as mixed evidence: entry failure with partially running core rules. Claude Opus 4.6 preserves the full board state, and the elimination loop remains observable.

\paragraph{Maze.} Both Kimi K2.5 and Claude Opus 4.6 can reach a visible victory state. This case serves mainly as a positive control: the runtime evaluator records not only failures, but can also follow the requirement-to-observation chain to verify path progress, player-position changes, and terminal feedback.

\paragraph{Lianliankan.} Kimi K2.5 is blocked during installation because its lockfile points to a private registry tarball. Under the scoring rule in Section~\ref{sec:experiments}, this is a valid \unusable{} delivery failure because the artifact cannot be entered by a user; in the case analysis, we separate it from browser-runtime interaction failures. Claude Opus 4.6 can load a playable board and preserves observable tile-removal state.

\paragraph{Blocktext.} Kimi K2.5's evidence package lacks intermediate interaction frames and retains only the menu or terminal screen. The more appropriate conclusion is therefore evidence missing, rather than direct runtime failure. Claude Opus 4.6 shows a visible main interface and game state, and its interaction loop can continue.

\section{Case Evidence for Runtime Evaluation}
\label{app:case_evidence}

The following cases illustrate the requirement-to-observation chain. They are not additional aggregate metrics.

\paragraph{Tetris-like game.} Human review noted that combo scoring and level progression are hard to verify through short natural play. The evaluator constructs candidate states for single-line clearing, four-line clearing, near-threshold score, spawn collision, and different tetromino types. After confirming each state, it triggers the final action through keyboard input. This process verifies line clearing and combo scoring, but exposes a level-progression bug: reaching the target score does not advance the level.

\paragraph{Farm management game.} The task contains multi-year decisions, resource constraints, random events, insufficient funds, and early termination under low sustainability. The evaluator constructs candidate states for insufficient funds, year-5 settlement, and low sustainability, and then submits the final real action. The main loop is playable, but the page lacks crop growth-state feedback and implements a technology event inconsistently with the specification.

\paragraph{Gold Miner game.} Natural play shows that the hook is too fast for reliable progress. The evaluator first verifies hook swing, capture, item value, and retrieval through real clicks, then constructs a level-clear candidate state to inspect the shop chain. Even after reaching the target score, the game still enters failure when the countdown ends; purchased power-ups also do not persist to the next level.

\paragraph{UNO-style game.} UNO prompts, draw behavior, function cards, and terminal conditions require specific hands. The evaluator uses natural play and candidate hands, then clicks visible cards or the draw pile. It finds that illegal plays can be accepted, emptying the player's hand does not immediately trigger settlement, and the visible draw-pile region is blocked by a decorative layer.

\end{document}